%% file: sample-sigconf.tex
\documentclass[sigconf]{acmart}

\usepackage{booktabs,multirow,array} 
\usepackage[ruled,linesnumbered]{algorithm2e}
\usepackage[english]{babel}
\usepackage{lipsum}
\usepackage{enumitem}
\usepackage{balance}
\usepackage{arydshln}
\usepackage{adjustbox}
\usepackage{bm}

\makeatletter
\def\BState{\State\hskip-\ALG@thistlm}
\makeatother

\copyrightyear{2022}
\acmYear{2022}
\setcopyright{acmcopyright}\acmConference[WWW '22]{Proceedings of the ACM Web Conference 2022}{April 25--29, 2022}{Virtual Event, Lyon, France}
\acmBooktitle{Proceedings of the ACM Web Conference 2022 (WWW '22), April 25--29, 2022, Virtual Event, Lyon, France}
\acmPrice{15.00}
\acmDOI{10.1145/3485447.3512020}
\acmISBN{978-1-4503-9096-5/22/04}

\begin{document}

\fancyhead{}

\title{User Satisfaction Estimation with Sequential Dialogue Act Modeling in Goal-oriented Conversational Systems}

\author{Yang Deng, Wenxuan Zhang, Wai Lam, Hong Cheng, Helen Meng}
\affiliation{%
  \institution{The Chinese University of Hong Kong}
  \country{Hong Kong SAR}
}
\email{{ydeng, wxzhang, wlam, hcheng, hmmeng}@se.cuhk.edu.hk}

\begin{abstract}
User Satisfaction Estimation (USE) is an important yet challenging task in  goal-oriented conversational systems. 
Whether the user is satisfied with the system largely depends on the fulfillment of the user's needs, which can be implicitly reflected by users' dialogue acts. However, existing studies often neglect the sequential transitions of dialogue act or rely heavily on annotated dialogue act labels when utilizing dialogue acts to facilitate USE. 
In this paper, we propose a novel framework, namely \textbf{USDA}, to incorporate the sequential dynamics of dialogue acts for predicting user satisfaction, by jointly learning \textbf{U}ser \textbf{S}atisfaction Estimation and \textbf{D}ialogue \textbf{A}ct Recognition tasks. 
In specific, we first employ a Hierarchical Transformer to encode the whole dialogue context, with two task-adaptive pre-training strategies to be a second-phase in-domain pre-training for enhancing the dialogue modeling ability. 
In terms of the availability of dialogue act labels, we further develop two variants of USDA to capture the dialogue act information in either supervised or unsupervised manners. 
Finally, USDA leverages the sequential transitions of both content and act features in the dialogue to predict the user satisfaction. 
Experimental results on four benchmark goal-oriented dialogue datasets across different applications show that the proposed method substantially and consistently outperforms existing methods on USE, and validate the important role of dialogue act sequences in USE. 
\end{abstract}
 
%
%



\maketitle

\input{samplebody-conf}

\bibliographystyle{ACM-Reference-Format}
\bibliography{sample-bibliography}

\input{appendix}

\end{document}

%% file: samplebody-conf.tex
\section{Introduction}
A variety of goal-oriented conversational systems have emerged for assisting users to automatically accomplish various goals, such as task-oriented dialogue~\cite{mwoz}, conversational recommendation~\cite{redial}, conversational information-seeking~\cite{cast}, etc. 
User Satisfaction Estimation (USE)~\cite{umap20,sigir21-uss,emnlp20-jointuse,naacl21-selfsuper-act,emnlp19-cusser-sat,ntcir15-dialeval} receives increasing attention as it attaches great importance in evaluating the performance of the dialogue systems as well as adjusting the system's strategy to better fulfill the user's goal. 
Recent studies in popular user-engaged web services, including search~\cite{wsdm15-search-sat,wsdm18-productintent-sat,wsdm19-searchintent-sat,sigir17-searchsat}, recommendation~\cite{www19-intentrec-sat}, and advertisement~\cite{cikm20-advintent-sat}, identify the interaction signals (\textit{e.g.}, browsing and clicking logs) and their temporal sequences (\textit{e.g.}, click streams) to be important features for estimating user satisfaction. 
Unfortunately, such explicit interaction signals are no longer available in dialogue systems where the users implicitly express their intents or provide feedback through natural language responses.

Earlier studies often treat USE in dialogue systems as a sentiment analysis~\cite{emnlp19-cusser-sat} or response quality assessment task~\cite{emnlp20-jointuse}, which predict the user satisfaction solely based on content features. 
However, similar to other user-engaged web services, whether the user is satisfied with the system during a goal-oriented conversation largely depends on whether the system is successful in meeting the user's needs~\cite{sigir21-uss}. 
Some researchers~\cite{umap20,sigir21-uss,aaai21-handoff} found that dialogue acts~\cite{cl20-da}, which represent the user intents or actions at each conversation turn, can well reflect the fulfillment of the user's goal. 
For instance, the statistics in \cite{umap20} show that the dialogue act ``\textit{Add Details}'' occurs more frequently in unsatisfactory conversational recommendation dialogues. 
Besides, it is also observed that users tend to take the action ``\textit{Contact Manual Service}'' when facing the system's failure in understanding their needs~\cite{sigir21-uss,aaai21-handoff,sigdial21-dissat}. 
Therefore, several attempts~\cite{sigdial09-dialog-sat,umap20,cikm18-dialintent-sat,sigir16-sat-intass} have been made on pipeline-based USE methods in goal-oriented conversational systems, where dialogue act recognition (DAR) is often viewed as an important preceding step of USE. 
\citet{cikm18-dialintent-sat} develop intent sensitive word embeddings for measuring user satisfaction, while \citet{umap20} classify user actions as features in a preceding step for USE.

\begin{figure}
\centering
\includegraphics[width=0.48\textwidth]{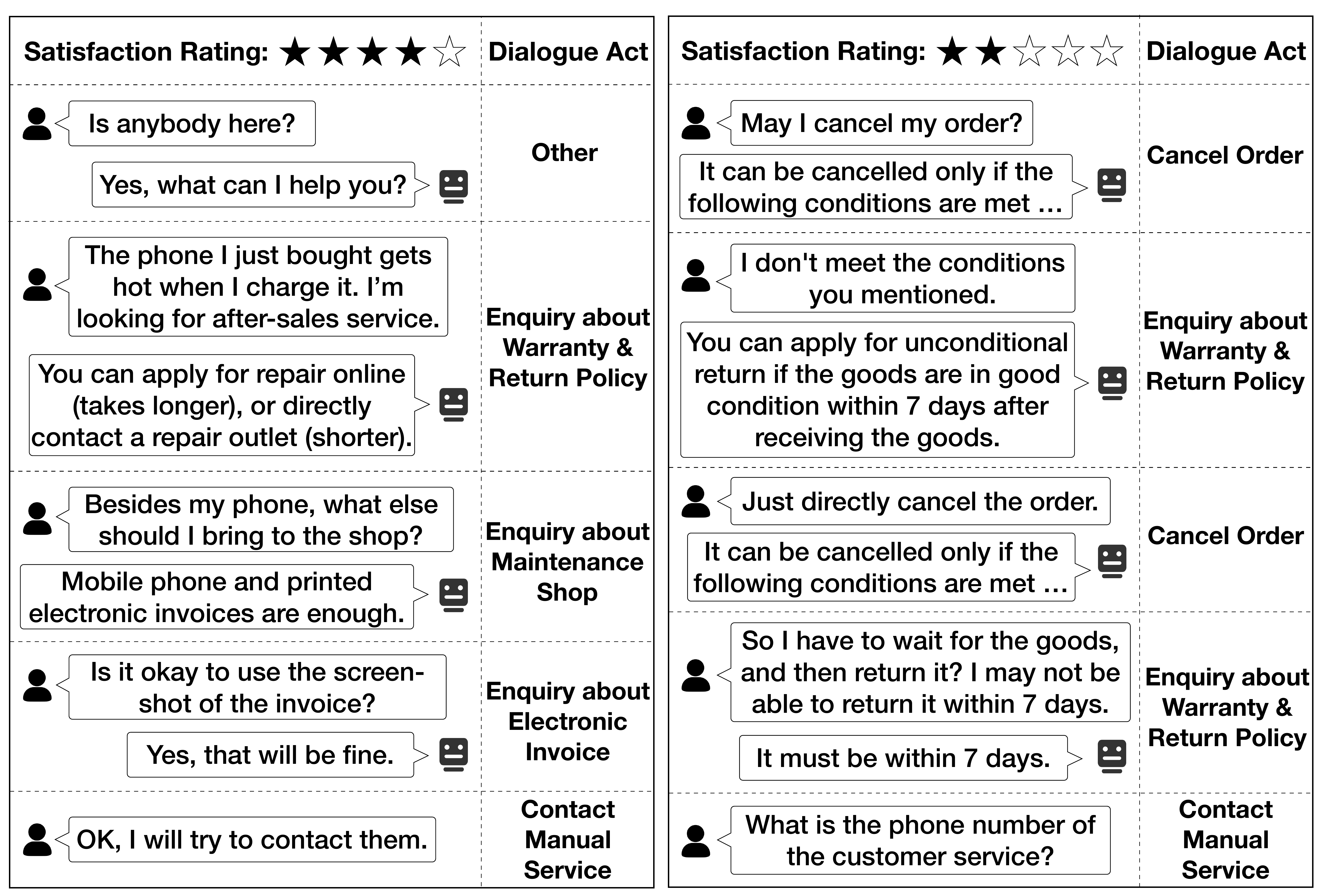}
\caption{Two example dialogue sessions in JDDC dataset~\cite{jddc}.}
\label{example}
\vspace{-0.3cm}
\end{figure}

Despite the effectiveness of incorporating dialogue acts on USE in previous works, there are several issues that remain to be tackled. 
(i) DAR~\cite{sigir18-crf-asn,cikm19-dar}, which aims to attach semantic labels to each utterance in the given dialogue for characterizing the speaker's intention, is still a challenging task in multi-turn dialogue systems. 
Pipeline-based approaches~\cite{umap20,cikm18-dialintent-sat}, which first detect the dialogue acts from user utterances and then adopt the predicted dialogue acts as the interaction features for USE, suffer severely from error propagation and inability to model interactions between the two tasks. 
(ii) The taxonomy for dialogue acts is diverse according to different application domains~\cite{sigir21-uss}, leading to a great expense for acquiring annotated dialogue act labels. 
(iii) Existing studies~\cite{umap20,sigir21-uss} initially investigate the relationship between user satisfaction and each individual dialogue act, while the sequential information behind the dialogue acts is neglected. 
As shown in Figure~\ref{example}, there are two dialogue sessions from a real-world E-Commerce customer service dialogue dataset. 
Although the final user acts in both dialogues are ``\textit{Contact Manual Service}'', they result in divergent satisfaction ratings. 
We can observe from the transition of user dialogue acts that the dialogue acts in the satisfactory dialogue (left) naturally transit to different topics, while some dialogue acts are repeated in the unsatisfactory dialogue (right).

To tackle the aforementioned issues, we propose a novel method, namely \textbf{USDA}, to jointly learn \textbf{U}ser \textbf{S}atisfaction Estimation and \textbf{D}ialogue \textbf{A}ct Recognition tasks. 
On one hand, DAR serves as an auxiliary task that provides clues about sequential user intents for USE. In return, the dialogue act transitions can also benefit from the prediction of user satisfaction. 
When the dialogue act labels are available, the joint learning aims at assigning a dialogue act label to each user utterance in the whole conversation session to represent her/his conversational intents, and meanwhile, estimating the degree of user satisfaction towards the conversation. 
When the dialogue act labels are unavailable, the DAR subtask is expected to be conducted in an unsupervised manner, which could still provide useful sequential patterns for helping estimate the user satisfaction. 

Specifically, we first develop a Hierarchical Transformer encoder, consisting of an exchange-level BERT encoder and a dialogue-level Transformer encoder, to encode the whole dialogue context. 
In terms of the availability of dialogue act labels, we design two variants of USDA to conduct either supervised DAR by a sequence labeling module or unsupervised DAR by a latent subspace clustering network. 
After modeling the user dialogue acts, a pair of attentive recurrent neural networks are employed to capture the sequential dynamics of dialogue transition from the perspective of both dialogue content and user acts. 
The sequential features of content and dialogue acts are ultimately aggregated by a gated attention mechanism to measure their importance in estimating user satisfaction. 
Overall, two tasks are jointly trained to leverage their interrelations in a multi-task learning procedure for mutually enhancing each other. 

In order to better model the dialogue context, we further propose two strategies for task-adaptive pre-training, namely System Response Selection (SRS) and Dialogue Incoherence Detection (DID), to serve as a second-phase in-domain pre-training for the Hierarchical Transformer encoder.
SRS endows USDA with the ability to identify whether the system's response is appropriate in the context of current conversation, which is essential to user satisfaction estimation. 
DID facilitates USDA in capturing the chronological semantic flow of dialogues by learning to detect sequential incoherence in dialogues, which can benefit the down-stream tasks when modeling the sequential features of the dialogue. 
Besides, they are conceptually simple, and the pre-training datasets can be created without any human annotation effort.

The contributions are summarized as follows:\vspace{-0.1cm}
\begin{itemize}[leftmargin=*]
    \item We propose to leverage the sequential dynamics of dialogue acts to facilitate USE in goal-oriented conversational systems via a unified joint learning framework, which enables both supervised and unsupervised DAR for the situations where the dialogue act annotations are available or not. 
    \item We introduce two task-adaptive self-supervised pre-training strategies, namely System Response Selection (SRS) and Dialogue Incoherence Detection (DID), for enhancing the dialogue modeling capability for helping the target joint learning problem. 
    \item Experimental results on four goal-oriented dialogue datasets show that USDA substantially and consistently outperforms existing methods on USE. Extensive analyses also reveal the relationships between content and dialogue act features for understanding user satisfaction in goal-oriented dialogue systems. 
\end{itemize}

\section{Related Work}
\textbf{Goal-oriented Conversational Systems.} 
Unlike chitchat-based dialogue systems~\cite{survey}, which aim at conversing with users on open-domain topics, task-oriented dialogue systems target at assisting users to accomplish certain goals~\cite{acl18-e2etod,mwoz,sgd}. 
In the broader area of dialogue systems, recent years have witnessed many other successful goal-oriented conversational applications. 
For instance, conversational recommender systems~\cite{redial,wsdm20-ear,umap20,sigir21-crs} enable traditional recommender systems to dynamically acquire user preferences through interactive conversations for making better recommendations. 
Conversational information-seeking platforms~\cite{chiir16-sat-intass,wise} play an increasingly important role in connecting users to search engines by conversational interactions. 
Customer service assistants provide timely helps on E-Commerce websites for answering product-related questions~\cite{tois22,cikm20-pqa} or handling after-sales problems~\cite{jddc}. 
In this paper, we conduct experiments across four different kinds of goal-oriented conversational systems to analyze user satisfaction. 

\textbf{User Satisfaction Estimation.} 
User satisfaction, which is related to the fulfillment of a specified desire or goal, is essential in evaluating and improving user-centered interactive IR systems~\cite{ftir09}. 
However, user satisfaction is difficult to measure automatically as it is rather subjective and implicit~\cite{sigir17-searchsat}. 
Therefore, several efforts have been made on modeling user satisfaction from temporal user behaviors or actions when interacting with the systems~\cite{sigir17-searchsat,wsdm18-productintent-sat,wsdm19-searchintent-sat,www19-intentrec-sat,cikm20-advintent-sat}. 
For instance, \citet{sigir17-searchsat} extract informative and interpretable action sequences (\textit{e.g.}, Click, Pause, Scroll) from user interaction data to predict user satisfaction towards the search systems.  
As for user satisfaction in dialogue systems~\cite{lrec10-dialog-sat,cikm18-dialintent-sat,sigdial09-dialog-sat}, the problem becomes more challenging since even the user behaviors and actions are hidden in the user's natural language feedback. 
Some researchers study user satisfaction in dialogue systems from the perspectives of sentiment analysis~\cite{emnlp19-cusser-sat} and response quality assessment~\cite{emnlp20-jointuse,ntcir15-dialeval}.  
Due to the nature of goal-oriented conversational systems, recent studies~\cite{umap20,sigir21-uss} identify the importance of dialogue acts or user intents in measuring the fulfillment of the user's goal, which is essential in user satisfaction estimation.   
However, the sequential dynamics of dialogue acts are neglected.

\textbf{Dialogue Act Recognition.} 
Dialogue Act Recognition (DAR)~\cite{sigir18-crf-asn,cikm19-dar,sigir19-CDAC} plays an important role in dialogue systems, as it reveals the user intents during the conversation session. 
Meanwhile, DAR often serves as an auxiliary task for improving other target conversation tasks, such as response generation~\cite{acl20-dar-generation} or selection~\cite{emnlp19-dar-mtrs,www20-intent-mtrs}, conversational information seeking~\cite{sigir18-intent-cis}, etc. 
With annotated dialogue act labels, DAR is typically formulated as a utterance-level sequence labeling task~\cite{sigir18-crf-asn,cikm19-dar,sigir19-CDAC}, which aims at capturing the sequential dependencies of the dialogue act sequence. 
However, the dialogue act label taxonomy heavily relies on domain-specific knowledge and varies from different domains, 
which inspires some unsupervised DAR studies~\cite{wsdm12-unsup-intent,emnlp19-intent-cluster,coling20-mtrs} for modeling user intents.

\begin{figure*}
\centering
\includegraphics[width=0.95\textwidth]{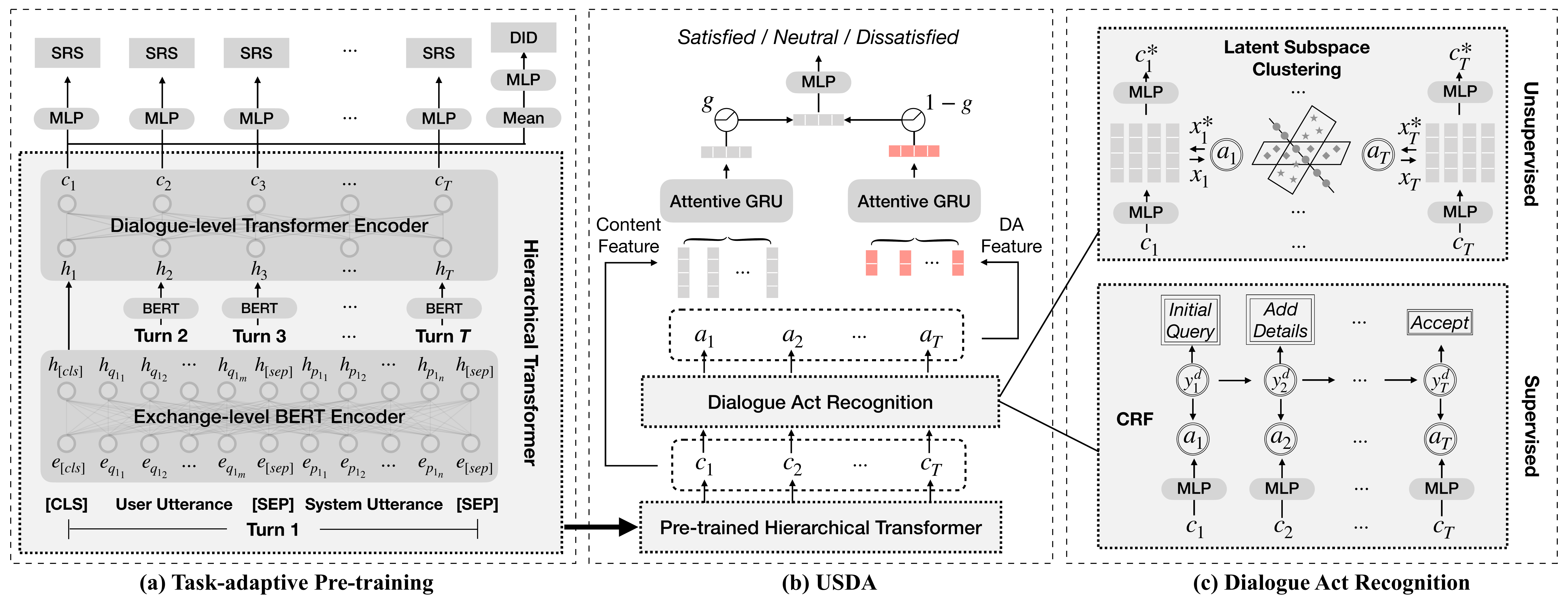}
\caption{Overview of USDA. USDA (MTL) and USDA (CLU) adopt supervised and unsupervised DAR modules respectively.}
\label{method}
\end{figure*}

\section{Problem Definition}
We define the joint learning of user satisfaction estimation (USE) and dialogue act recognition (DAR) tasks. 
Given the dialogue context $\mathcal{D}$ containing $N$ utterances $\{u_1,u_2,...,u_N\}$ in a conversation session, we split the dialogue into $T$ exchanges, and each exchange is a conversation turn between the user and the system, so that $\mathcal{D}$ can be represented by  $\{(u_{q_1},u_{p_1}),(u_{q_2},u_{p_2}),...,(u_{q_{T-1}},u_{p_{T-1}}),(u_{q_{T}},)\}$, where $u_{q_t}$ and $u_{p_t}$ denote the user utterance and the system utterance at $t$-th exchange. 
Note that the $T$-th exchange only contains the final user response. 
The goal is to simultaneously predict the sequence of dialogue act (DA) labels $\{y^d_{1},y^d_{2},...,y^d_{T}\}$  corresponding to the user intent at each turn, and the user satisfaction label $y^s$ by the end of this conversation session. 
As for DAR task, it can be accomplished via either supervised learning with annotated DA labels or unsupervised learning as an DA clustering task.  

\section{Method}\label{sec:method}
The overview of the proposed framework, USDA, is depicted in Figure~\ref{method}(b). 
There are two variants of USDA, including (1) USDA (MTL) conducts the multi-task learning of USE and DAR with the ground-truth DA labels. (2) USDA (CLU) jointly learns USE and unsupervised DAR with a latent subspace clustering module for applications where the DA labels are unavailable.  

\subsection{Hierarchical Transformer Encoder}
Pre-trained language models, like BERT~\cite{bert}, have become the de-facto method for encoding natural language into distributed representations in a wide range of natural language understanding tasks, including USE~\cite{sigir21-uss} and DAR~\cite{aaai21-cogat}. 
We also leverage BERT as the backbone of the proposed method. 
However, the target problem involves two tasks at different levels of dialogues, while the original BERT could only take the whole dialogue as input and suffers from its length limitation (512 tokens), which may largely harm the performance.   
To this end, we develop a hierarchical Transformer encoder for the representation learning of dialogue context. 
Such encoder consists of a shared Exchange-level BERT Encoder for pairwisely encoding the exchange between the system and user in each turn, and a Dialogue-level Transformer Encoder for sequentially encoding the dialogue context in the whole conversation session. 

\subsubsection{Exchange-level BERT Encoder}
Each exchange turn $(u_{q_t},u_{p_t})$ is first fed into a shared BERT encoder to obtain the exchange-level representation $h_t$:
\begin{equation}
    h_t = \mathbf{BERT}([CLS];u_{q_t};[SEP];u_{p_t};[SEP]),
\end{equation}
which also takes advantage of the  pairwise-learning capability of BERT~\cite{bert} to model the relation between utterances. 

\subsubsection{Dialogue-level Transformer Encoder}
To capture the global context information in the whole conversation session, we employ a Transformer encoder on top of the exchange-level BERT encoder. 
The dialogue-level Transformer encoder takes the representations $\{h_1,h_2,...,h_T\}$ of all the exchanges as input. 
Following the standard Transformer encoder~\cite{transformer}, we also add positional encoding to the exchange representations for modeling the relative positions of the exchange in the whole conversation. 
Each Transformer encoder layer consists of three components: (i) The layer normalization is defined as LayerNorm$(\cdot)$. (ii) The multi-head attention is defined as MultiHead$(\bm{Q}, \bm{K}, \bm{V})$, where $\bm{Q}, \bm{K}, \bm{V}$ are query, key, and value, respectively. (iii) The feed-forward network with ReLU activation is defined as FFN$(\cdot)$. Take the $l$-th layer for example: 
\begin{align}
    \bm{X}^* &= \text{MultiHead}(\bm{X}^{(l)},\bm{X}^{(l)},\bm{X}^{(l)}), \\
    \bm{X}^{(l+1)} &=\text{LayerNorm}(\text{FFN}(\bm{X}^*) + \bm{X}^{(l)}),
\end{align}
where $X^{(0)}=\{h_1,h_2,...,h_T\}$ is the input of the first layer, and $X^{(L)}=\{c_1,c_2,...,c_T\}$ is the output of the last layer. 
$c_t$ is the contextualized representation for the exchange at $t$-th turn.

\subsection{Dialogue Act Recognition}
Due to the difficulties in obtaining the DA taxonomy in a new domain, we develop two variants of the DAR module for the joint learning. As shown in Figure~\ref{method}(c), the DAR task can be accomplished by either \textbf{supervised learning} with ground-truth DA labels, \textit{i.e.}, USDA (MTL), or \textbf{unsupervised learning} into latent DA clusters, \textit{i.e.}, USDA (CLU), in terms of the availability of the DA labels. 

\subsubsection{Supervised Learning}
With the contextualized representations $\{c_1,c_2,...,c_T\}$ of each exchange, we can directly calculate the probability distribution of dialogue act labels for the user utterance at each exchange turn by a Multi-Layer Perceptron (MLP):  
\begin{align}
    a_t &=\mathbf{MLP}_\text{DAR}(c_t),
\end{align}
where $A=\{a_1,a_2,...,a_T\}\in \mathbb{R}^{T\times K}$ is adopted as the dialogue act score, where $K$ is the number of DA classes. Then, a Conditional Random Field (CRF)~\cite{crf} layer is used to capture the dependencies among consecutive DA labels, which models the conditional probability of a target label sequence given an input sequence. 
Let $A_{t,y^d_{t}}$ denote the dialogue act score of the dialogue act $y^d_{t}$ of the $t$-th exchange turn in a conversation. The DA sequence scores $\sigma(A,y^d)$ and likelihood $p(y^d|A)$ are calculated as follows:
\begin{align}
    \sigma(A,y^d)&=\sum\nolimits_{t=0}^{T+1} \bm{G}_{y^d_t,y^d_{t+1}} +\sum\nolimits_{t=1}^T A_{t,y^d_{t}},\\
    p(y^d|A) &= \text{softmax}(\sigma(A,y^d)),
\end{align}
where $\bm{G}\in\mathbb{R}^{(K+2)\times (K+2)}$ is a matrix of transition scores such that $\bm{G}_{i,j}$ represents the score of a transition from the
$i$-th DA label to $j$-th DA label. 
$y^d_0$ and $y^d_{T+1}$ are the start and end labels of a dialogue. 
During training, the supervised learning objective is to minimize the negative log-probability of the correct dialogue act sequence:
\begin{equation}
    \mathcal{L}_\text{DAR}^{+} = -\log p(y^d|A).
\end{equation}

\subsubsection{Unsupervised Learning}
Under the circumstance where the dialogue act labels are unavailable, previous works~\cite{cikm18-dialintent-sat,emnlp19-intent-cluster,coling20-mtrs} induce dialog intents by clustering user utterances to learn discriminative utterance representations in the user intent semantic space. 
Since the user utterances are encoded into high-dimensional contextualized vectors, we employ a deep latent subspace clustering network~\cite{nips17-clustering} for the representational learning of latent features in the dialog to capture the DA information. 
The basic idea is to encourage utterances of the same dialogue act to be clustered into a dense region in the low-dimensional embedding subspace. 

We initialize the latent dialogue act memory vectors $M\in\mathbb{R}^{K\times d}$, where $K$ is the number of dialogue act clusters, and measure the similarity between the sentence representation and each latent memory vector via a deep  auto-encoder. 
In specific, the contextualized exchange representations $C=\{c_1,c_2,...,c_T\}$ are first encoded into the latent representations $X$ by an MLP. 
Then the latent representations are self-represented
by a self-attentive weighted sum of the latent clustering memory vectors. 
After the self-representation operation, the contextualized representations $C^*$ are reconstructed by another MLP as the decoder. 
\begin{align}
    X &= \mathbf{MLP}_{\text{enc}}(C),\quad A = X M^\top,\\
    X^* &= \text{softmax}(A) M,\quad C^* = \mathbf{MLP}_{\text{dec}}(X^*),
\end{align}
where $A=\{a_1,a_2,...,a_T\}$ represents the dialogue act features learned from the deep latent subspace clustering network. 

The loss function of the latent dialogue act representation learning consists of three parts, information preservation loss,  self-representation loss, and a regularization term:
\begin{equation}
    \mathcal{L}_\text{DAR}^- = \underbrace{||C^*-C||^2_F}_\text{info. preservation} + \underbrace{\lambda_1||X^*-X||^2_F}_\text{self-representation} + \underbrace{\lambda_2||M M^\top - I||}_\text{regularization},
\end{equation}
where $\lambda_1$ and $\lambda_2$ are the hyper-parameters that balance the weight of different terms, and $I$ is the identity matrix. The information preservation loss ensures that the information from the contextual representation is encoded into the latent representation, while the self-representation loss aims to minimize the differences between the common clustering representation and the latent representation. 
The regularization term is to differentiate the memory vector for each dialogue act cluster from other clusters.

\subsection{User Satisfaction Estimation}
With the content features $c_t$ learned from the hierarchical transformer encoder and the dialogue act features $a_t$ learned from the DAR task, we employ a pair of Attentive RNNs to capture both the content-based and action-based dialogue transitions within a conversation session: 
\begin{equation}
    v^c_t = \mathbf{GRU}_c([c_t],v^c_{t-1}),\quad v^a_t = \mathbf{GRU}_a([a_t],v^a_{t-1}),
\end{equation}
where $V_c=\{v^c_1,...,v^c_T\}$ and $V_a=\{v^a_1,...,v^a_T\}$ are the output representations for content features and DA features, respectively.  

Then we employ a vanilla attention mechanism to attend those important exchange turns for predicting user satisfaction from  the perspectives of both content and dialogue act:
\begin{align}
    \alpha_c &= \text{softmax}(w_c^\top\text{tanh}(V_cW_c^\top)),\quad o_c = V_c^\top\alpha_c,\\
    \alpha_a &= \text{softmax}(w_a^\top\text{tanh}(V_aW_a^\top)),\quad o_a = V_a^\top\alpha_a,\label{eq:att}
\end{align}
where $W_c, W_a, w_c, w_a$ are trainable parameters. 
An aggregation layer integrates the attentive content representations $o_c$ and DA representations $o_a$ by the following gated attention mechanism:
\begin{align}
    g &= \text{sigmoid}(W_g[o_c;o_a]),\label{eq:gate}\\
    o &= g*o_a + (1-g)*o_c,
\end{align}
where $W_g$ is the attention matrix to be learned, and $g$ is the gated attention weight. Then we can calculate the probability distribution of user satisfaction classes by an MLP with softmax normalization: 
\begin{equation}
    p^\text{USE}=\text{softmax}(\mathbf{MLP}_\text{USE}(o)).
\end{equation}

\subsection{Joint Learning Procedure}
USDA jointly learns USE and DAR in an end-to-end fashion. 
As for DAR, we can choose to either use the supervised learning or unsupervised learning objectives, \textit{i.e.}, $\mathcal{L}_\text{DAR}^{+}$ or $\mathcal{L}_\text{DAR}^{-}$, in terms of whether the dialogue act labels are available. 
And USE objective is to minimize the cross-entropy loss between the predicted user satisfaction probabilities and the ground-truth satisfaction classes:
\begin{equation}
    \mathcal{L}_\text{USE} = -y^s\log p^\text{USE}. 
\end{equation}
Finally, the overall loss function is the sum of USE and DAR losses: 
\begin{equation}
    \mathcal{L} = \mathcal{L}_\text{USE} + \lambda\mathcal{L}_\text{DAR}^\dagger, \quad \dagger\in [+, -], 
\end{equation}
where $\lambda$ is the hyper-parameter balancing the two tasks.

\subsection{Task-adaptive Pre-training Strategies}\label{sec:pretrain}
The target problem involves two tasks at different levels of dialogues, which requires the capability of both contextual modeling and sequential modeling. 
Appropriate continued pre-training on the target tasks or domains can further enhance the capability of representational learning in pre-trained language models~\cite{acl20-dontstop}. 
Motivated by this, we customize two auxiliary self-supervised pre-training tasks to serve as a second-phase in-domain pre-training for the Hierarchical Transformer without any extra human annotation. 

\subsubsection{System Response Selection (SRS)}
Response selection is an extensively studied task in retrieval-based dialogue systems~\cite{www20-intent-mtrs}, which aims at selecting appropriate utterances from a set of candidates to respond to the given dialogue context. 
As for USE task, the appropriateness or relevancy of the system-provided response is also highly related to the user satisfaction. 
Therefore, our USDA model is expected to be capable of identifying whether a system-provided response is appropriate to the given dialogue context. 
In practice, we regard the ``satisfactory'' dialogues in the original training dataset as the positive samples for SRS task. 
To construct the negative sample, we randomly choose a different utterance  from the rest of the dialogues to replace one of the system responses in the original ``satisfactory'' dialogue, which means that the replaced system response is inappropriate at the corresponding turn.

\subsubsection{Dialogue Incoherence Detection (DID)}
Detecting incoherence in narratives~\cite{nid} is an important ability in natural language understanding, which is the same in dialogue systems~\cite{acl20-dar-did}. 
On one hand, it is required to model the sequential information in the dialogue context for DAR task. 
On the other hand, whether the user is satisfied is also highly related to the dialogue coherency. 
Motivated by this, we expect that our USDA model can also detect the incoherence in the given dialogue. 
The positive samples are the same as those in SRS task, while we employ two ways to construct the incoherent dialogues as the negative samples:  
(i) We randomly delete several exchanges in the original dialogue as the negative sample for missing information detection. 
(ii) We randomly shuffle several exchanges in the original dialogue for discordant dialogue detection.
DID is a dialogue-level classification task to determine whether the dialogue is discordant or missing information.

\subsubsection{Joint Task-adaptive Pre-training}
Similar to the joint learning procedure, we conduct the joint task-adaptive pre-training of SRS and DID tasks for the overall enhancement of the Hierarchical Transformer encoder, which is depicted in Figure~\ref{method}(a). 
As for the constructed SRS samples, the negative samples can also be regarded as incoherent dialogues. 
As for the constructed DID samples, all the system responses are supposed to be appropriate, since each individual exchange turn is unchanged. 
More details about the task-adaptive pre-training can be found in Appendix~\ref{apx:pretrain}.

\section{Experimental Setups}\label{sec:exp_setup}

\begin{table}
    \caption{Satistics of datasets.}
    \centering
    \begin{tabular}{lrrrr}
    \toprule
        Dataset & MWOZ & SGD&JDDC&ReDial \\
        \midrule
        Language& English & English & Chinese & English\\
        \#Dialogues&1,000&1,000&3,300&1,000\\
        \#Utterances&12,553&13,833&54,517&11,806\\
        \%Sat. Class&27:39:34&22:30:48&23:53:24&23:26:51\\
        \#DA Classes&21&12&236&17\\
        \#Avg Turn&8.66&9.42&10.93&8.49\\
        \midrule
        \#TrainSplit&7,648&8,674&38,146&7,372\\
        \#ValidSplit&952&1,074&5,006&700\\
        \#TestSplit&953&1,085&4,765&547\\
    \bottomrule
    \end{tabular}
    \label{dataset}
    \vspace{-0.3cm}
\end{table}

\subsection{Datasets \& Evaluation Metrics}
We evaluate the proposed method on four benchmark task-oriented dialogue datasets varying from different applications, including MultiWOZ2.1 (MWOZ)~\cite{mwoz}, Schema Guided Dialogue (SGD)~\cite{sgd}, JDDC~\cite{jddc}, and ReDial~\cite{redial}. 
In specific, we adopt the subsets of these datasets with the user satisfaction annotation for evaluation, which is provided by \citet{sigir21-uss}: 
\begin{itemize}[leftmargin=*,topsep=4pt]
    \item \textbf{MWOZ} and \textbf{SGD} are multi-domain task-oriented dialogue datasets with dialogue act annotation. 
    \item \textbf{JDDC} is a Chinese customer service dialogue dataset in E-Com-merce with DA labels annotated by \citet{sigir21-uss}. 
    \item \textbf{ReDial} is a conversational recommendation dataset for movie recommendation. The DA label taxonomy is annotated by \citet{umap20}. Since there are only a small number of overlapping instances between the annotated subsets in \cite{sigir21-uss} and those in \cite{umap20}, we adopt those overlapping instances as the testing set while the rest of annotated data in \cite{sigir21-uss} as the training and validation set for unsupervised dialogue act recognition. 
\end{itemize}

\begin{table*}
\centering
\caption{Method comparisons on MWOZ, SGD, and JDDC datasets.  $^\dagger$ indicates that the model is better than the best performance of baseline methods (\underline{underline} scores) with statistical significance (measured by significance test at $p<0.05$).}
\begin{adjustbox}{max width=\textwidth}
\setlength{\tabcolsep}{0.8mm}{
\begin{tabular}{l:cccccccc:cccccccc:cccccccc}
\toprule
\multicolumn{1}{c}{}&\multicolumn{8}{c}{MWOZ}& \multicolumn{8}{c}{SGD}& \multicolumn{8}{c}{JDDC}\\ 
\cmidrule(lr){2-9}\cmidrule(lr){10-17}\cmidrule(lr){18-25}
\multicolumn{1}{c}{Model}&\multicolumn{4}{c}{USE}&\multicolumn{4}{c}{DAR}& \multicolumn{4}{c}{USE}&\multicolumn{4}{c}{DAR}&\multicolumn{4}{c}{USE}&\multicolumn{4}{c}{DAR}\\
\cmidrule(lr){2-5}\cmidrule(lr){6-9}\cmidrule(lr){10-13}\cmidrule(lr){14-17}\cmidrule(lr){18-21}\cmidrule(lr){22-25}
\multicolumn{1}{c}{}&Acc&P&R&F1&Acc&P&R&\multicolumn{1}{c}{F1}&Acc&P&R&F1&Acc&P&R&\multicolumn{1}{c}{F1}&Acc&P&R&F1&Acc&P&R&\multicolumn{1}{c}{F1}\\
\midrule
\multicolumn{9}{l:}{\textit{Single-task Learning Methods}}&\multicolumn{8}{l:}{}\\
CRF-ASN~\cite{sigir18-crf-asn}&-&-&-&-&79.5&71.4&72.1&70.0&-&-&-&-&88.9&82.6&83.6&82.2&-&-&-&-&66.1&45.4&46.0&44.1\\
CRNN~\cite{cikm19-dar}&-&-&-&-&\underline{80.9}&\underline{73.4}&\underline{73.5}&\underline{71.9}&-&-&-&-&\underline{90.2}&\underline{85.3}&\underline{85.9}&\underline{84.7}&-&-&-&-&\underline{66.4}&\underline{45.7}&\underline{46.6}&\underline{44.6}\\
HiGRU~\cite{naacl19-higru}&44.6&43.7&44.3&43.7&-&-&-&-&50.0&47.3&48.4&47.5&-&-&-&-&59.7&57.3&50.4&52.0&-&-&-&-\\
HAN~\cite{sigir21-uss}&39.0&37.1&37.1&36.8&-&-&-&-&47.7&47.1&44.8&44.9&-&-&-&-&58.4&54.2&50.1&51.2&-&-&-&-\\
BERT~\cite{bert}&\underline{46.1}&\underline{45.5}&\underline{47.4}&\underline{45.9}&-&-&-&-&\underline{56.2}&\underline{55.0}&\underline{53.7}&\underline{53.7}&-&-&-&-&\underline{60.4}&\underline{59.8}&\underline{58.8}&\underline{59.5}&-&-&-&-\\
\hdashline
\multicolumn{1}{l}{\textbf{USDA (STL)}}&\textbf{49.9}$^\dagger$&\textbf{49.2}$^\dagger$&\textbf{49.0}$^\dagger$&\textbf{48.9}$^\dagger$&\textbf{87.7}$^\dagger$&\textbf{82.8}$^\dagger$&\textbf{82.3}$^\dagger$&\multicolumn{1}{c}{\textbf{81.3}$^\dagger$}&\textbf{61.4}$^\dagger$&\textbf{60.1}$^\dagger$&\textbf{55.7}$^\dagger$&\textbf{57.0}$^\dagger$&\textbf{95.2}$^\dagger$&\textbf{92.5}$^\dagger$&\textbf{92.3}$^\dagger$&\multicolumn{1}{c}{\textbf{92.0}$^\dagger$}&\textbf{61.8}$^\dagger$&\textbf{62.8}$^\dagger$&\textbf{63.7}$^\dagger$&\textbf{61.7}$^\dagger$&\textbf{69.4}$^\dagger$&\textbf{52.0}$^\dagger$&\textbf{52.0}$^\dagger$&\textbf{50.3}$^\dagger$\\
\midrule
\midrule
\multicolumn{25}{l}{\textit{Multi-task Learning Methods}}\\
JointDAS~\cite{jointdas}&44.8&42.7&43.0&42.8&75.1&64.5&64.7&62.8&55.7&52.2&52.4&52.3&79.5&72.1&72.7&70.9&58.5&55.8&55.1&55.4&63.4&41.8&43.6&41.1\\
Co-GAT~\cite{aaai21-cogat}&46.8&44.8&44.0&44.2&75.6&68.5&68.4&66.6&56.8&55.9&55.9&55.6&87.5&80.9&81.5&80.2&60.2&59.3&62.9&60.1&64.2&42.5&43.6&41.5\\
~~ + BERT&47.0&46.4&47.2&46.3&\underline{86.2}&\underline{79.8}&\underline{80.1}&\underline{78.8}&58.6&55.2&55.7&55.5&\underline{92.5}&88.2&88.3&87.6&60.6&60.6&\underline{63.7}&\underline{61.0}&66.7&\underline{49.4}&\underline{48.9}&\underline{47.5}\\
JointUSE~\cite{emnlp20-jointuse}&47.6&44.6&44.9&44.7&76.5&68.7&67.7&66.9&57.4&55.0&54.8&54.7&85.0&78.0&78.9&77.3&58.3&56.6&58.7&57.2&61.8&39.0&41.8&38.8\\
~~ + BERT&\underline{48.9}&\underline{47.2}&\underline{48.0}&\underline{47.3}&84.4&77.4&78.0&76.3&\underline{59.0}&\underline{57.4}&\underline{57.1}&\underline{57.3}&92.4&\underline{88.3}&\underline{88.5}&\underline{87.7}&\underline{63.8}&\underline{60.8}&58.6&59.2&\underline{66.8}&49.2&48.7&47.3\\
\hdashline
\textbf{USDA (CLU)}&\textbf{53.4}$^\dagger$&50.8$^\dagger$&49.9$^\dagger$&50.2$^\dagger$&-&-&-&-&61.6$^\dagger$&58.2$^\dagger$&59.3$^\dagger$&58.6$^\dagger$&-&-&-&-&\textbf{65.1}$^\dagger$&\textbf{63.5}$^\dagger$&\textbf{64.9}$^\dagger$&\textbf{64.0}$^\dagger$&-&-&-&-\\
\textbf{USDA (MTL)}&52.9$^\dagger$&\textbf{51.8}$^\dagger$&\textbf{50.2}$^\dagger$&\textbf{50.6}$^\dagger$&\textbf{87.7}$^\dagger$&\textbf{82.8}$^\dagger$&\textbf{82.4}$^\dagger$&\textbf{81.4}$^\dagger$&\textbf{62.5}$^\dagger$&\textbf{60.3}$^\dagger$&\textbf{59.9}$^\dagger$&\textbf{60.1}$^\dagger$&\textbf{95.8}$^\dagger$&\textbf{93.6}$^\dagger$&\textbf{93.4}$^\dagger$&\textbf{93.1}$^\dagger$&63.0&61.4&65.7$^\dagger$&62.6$^\dagger$&\textbf{69.7}$^\dagger$&\textbf{53.1}$^\dagger$&\textbf{53.0}$^\dagger$&\textbf{51.3}$^\dagger$\\
\bottomrule
\end{tabular}}
\end{adjustbox}
\label{result}
\vspace{-0.2cm}
\end{table*}

There are two typical settings for measuring user satisfaction: (i) five-level ratings (1-5) to distinguish the degree of user satisfaction from ``very dissatisfied'' to ``very satisfied''~\cite{emnlp20-jointuse,sigir21-uss}, (ii) three satisfaction classes to indicate the polarity of user satisfaction, \textit{i.e.}, ``dissatisfied/neural/satisfied''~\cite{emnlp19-cusser-sat,umap20,cikm19-offon-sat}. 
Since the boundaries between ``very dis/satisfied'' and ``dis/satisfied'' are often blurred and it is also more practical to classify the polarity of user satisfaction in real-world applications~\cite{sigdial21-dissat,aaai21-handoff}, we follow the second setting to study a three-class classification task for USE, and treat the average rating ``</=/> 3'' as ``dissatisfied/neural/satisfied''. 
Due to the absence of data split in the original dataset~\cite{sigir21-uss}, we split each dataset by 8:1:1 for train-valid-test set and filter out the dialogues with less than 2 turns. 
Table~\ref{dataset} presents the statistics of these four datasets\footnote{Available at \url{https://github.com/dengyang17/USDA}.}. 

Following previous studies~\cite{emnlp19-cusser-sat,umap20,cikm19-offon-sat}, we use Accuracy (Acc) and Macro-averaged Precision (P), Recall (R), and F1 as the evaluation metrics for both USE\footnote{We do not use Correlation metrics (\textit{e.g.}, Spearman or Pearson) for USE evaluation in this work, since they are typically adopted for the setting of five-level ratings.} and DAR tasks.

\subsection{Compared Methods}
We first compare the proposed method to several state-of-the-art \textit{single-task learning} methods on both DAR and USE tasks. 
\begin{itemize}[leftmargin=*,topsep=4pt]
    \item CRF-ASN~\cite{sigir18-crf-asn} extends structured attention network to the linear-chain conditional random field for DAR.
    \item CRNN~\cite{cikm19-dar} is a Convolutional Recurrent Neural Network modeling the interactions between utterances of long-range context.
    \item HiGRU~\cite{naacl19-higru} uses a hierarchical GRU structure to encode the dialogue context representation. 
    \item HAN~\cite{naacl16-hiattgru} applies a two-level attention mechanism in HiGRU.
    \item BERT~\cite{bert} concatenates the last 512 tokens of the dialogue context into a long sequence with a $[SEP]$ token for separating utterances. For HiGRU, HAN, and BERT, we adopt the implementation from \cite{sigir21-uss} for USE task.
\end{itemize}

To investigate the \textit{multi-task learning} of USE and DAR task, we adopt several alternative multi-task learning models from related tasks for comparisons. 
\begin{itemize}[leftmargin=*,topsep=4pt]
    \item JointDAS~\cite{jointdas} and Co-GAT~\cite{aaai21-cogat} are two state-of-the-art methods to jointly perform DAR and sentiment classification task for each utterance in the given dialogue context. In our case, we adopt the classification result of the last user utterance for USE task. 
    \item JointUSE~\cite{emnlp20-jointuse} is an USE approach which jointly predicts turn-level response quality labels provided by experts and dialogue-level ratings provided by end users. In our case, we replace the turn-level response quality labels with the dialogue act labels for DAR task, and omit the hand-crafted input features. 
\end{itemize}

For the proposed \textbf{USDA} method, we report the performance of three variants as follows:
\begin{itemize}[leftmargin=*,topsep=4pt]
    \item \textbf{USDA (STL)} performs solely USE or DAR task by discarding the DAR or the USE components. 
    \item \textbf{USDA (CLU)} jointly learns USE and unsupervised DAR  where we assume that the ground-truth DA labels are unavailable.
    \item  \textbf{USDA (MTL)} conducts the multi-task learning of USE and DAR with the ground-truth DA labels.
\end{itemize}

For a fair comparison and without loss of generality, we adopt BERT as the backbone for all methods with pre-trained language models. The implementation details can be found in Appendix~\ref{apx:impl}.

\begin{table*}
    \centering
    \caption{List of inferred dialogue acts for ReDial dataset, with top representative words and examples.}
    \label{cluster}
    \begin{adjustbox}{max width=\textwidth}
    \begin{tabular}{lll}
    \toprule
    Inferred Dialogue Acts&Representative Words&Examples\\
    \midrule
    Initial Query/Start Over&looking, interested, recently, want, please, today&I'm interested in war movies. Any good suggestions? \\
    Reformulate/Continue&mostly, especially, first, any, another, other, suggest&Do you have any other recommendations?\\
    Inquire/Ask Opinion&who’s, what’s, did’t, when, end, done&I haven't heard of it. What's it about? \\
    Accept/Provide Preference&everyone, rather, keep, big, better, best, interesting&That one sounds really interesting\\
    Reject/Critique-Feature&seem, isn’t, don’t, quite, couldn’t, can’t, wasn’t&I don't care for  "Grease  (1978)" ; not my type of movie.\\
    Seen&year, too, yes, have, seen, like, liked, great&I have. I liked it a lot. That was a great one.\\
    \bottomrule
    \end{tabular}
    \end{adjustbox}
\end{table*}

\begin{table}
\fontsize{8.5}{10}\selectfont
    \centering
    \caption{Experimental results on ReDial dataset. }
    \label{unsupervised}
    \begin{tabular}{lcccc}
    \toprule
    Model&Acc&P&R&F1\\
    \midrule
    HiGRU~\cite{naacl19-higru}&46.1&44.4&44.0&43.5\\
    HAN~\cite{naacl16-hiattgru}&46.3&40.0&40.3&40.0\\
    BERT~\cite{bert}&\underline{53.6}&\underline{50.5}&\underline{51.3}&\underline{50.0}\\
    \midrule
    \textbf{USDA (STL)}&57.3$^\dagger$&54.3$^\dagger$&52.9$^\dagger$&53.4$^\dagger$\\
    \textbf{USDA (CLU)}&\textbf{59.0}$^\dagger$&\textbf{55.5}$^\dagger$&\textbf{55.5}$^\dagger$&\textbf{55.5}$^\dagger$\\
    \bottomrule
    \end{tabular}
    \vspace{-0.2cm}
\end{table}

\section{Experimental Results}\label{sec:exp_result}

\subsection{Overall Performance}\label{sec:overall}
\subsubsection{Method Comparisons on MWOZ, SGD, and JDDC} 
Table~\ref{result} summarizes the experimental results on MWOZ, SGD, and JDDC datasets. 
Since these datasets contain the annotated DA labels for supervised DAR learning, we conduct extensive method comparisons, including single-task learning and multi-task learning. 
There are several notable observations as follows: 

(1) The proposed method substantially and consistently outperforms all the strong baselines across three datasets with a noticeable margin, including both single-task and multi-task learning methods. The results in single-task setting demonstrate the effectiveness of the Hierarchical Transformer encoder for dialogue modeling than the original BERT encoder.

(2) Compared with single-task learning, multi-task learning methods generally achieve better performance on USE task, which validates that USE can actually benefit from jointly learning with DAR. For instance, USDA (MTL) outperforms USDA (STL) on MWOZ and SGD datasets by about 4\%.
  
(3) Even when the DA labels are not provided, USDA (CLU) effectively utilizes the latent user intent to improve the performance of USE task via the proposed latent subspace clustering method. 
It is worth noting that USDA (CLU) performs much better than USDA (STL) on JDDC dataset. It can be observed that the DAR performance is relatively low on JDDC dataset (F1 around 50\%), as it conducts a 236-class classification task given the DA taxonomy. Therefore, the limited performance on DAR task hinders the improvement on USE task under the multi-task learning setting.

\subsubsection{Method Comparisons on ReDial} 

For ReDial dataset, where the DA labels are not provided, we conduct the comparisons under the single-task learning settings for USE.   
As shown in Table~\ref{unsupervised}, it can be observed that the hierarchical dialogue modeling structure (USDA (STL)) still outperforms the original BERT. 
Besides, the proposed latent subspace clustering of user intents, \textit{i.e.}, USDA(CLU), contributes a great performance boosting, about 5\% on F1 score. 
To facilitate further investigation of the latent subspace clustering, we derive the probability of words in each cluster, and rank them by their frequency. After filtering the stop words, the results of clusters and words from ReDial Dataset are presented in Table~\ref{cluster}. 
Compared with the DA taxonomy provided by \citet{umap20}, USDA (CLU) effectively differentiates user intents into different DA clusters in an unsupervised manner.

\subsubsection{Summary}
Overall, the above results demonstrate the effectiveness and the strong applicability of the proposed USDA methods. On one hand, given the ground-truth DA labels, USDA (MTL) can improve the USE performance via multi-task learning with DAR task. On the other hand, USDA (CLU) can also leverage latent user intents to aid in the USE task via the latent subspace clustering, when the DA labels are unavailable or the DAR task is relatively difficult. 
In addition, we also conduct analyses of per-class performance and the effect of the dialogue turn number, which are presented in Appendix~\ref{apx:disc}. 
It can be observed that most of the performance improvement of USDA comes from the ``dissatisfied'' class, which is the most important class for USE in real-world applications~\cite{aaai21-handoff,sigdial21-dissat}. 
And USDA can reach a better performance when involving a certain number of conversation turns, instead of decreasing significantly.

\subsection{Ablation Study}
\subsubsection{Effect of Components}
To understand the importance of DA features for predicting user satisfaction in goal-oriented conversational systems, we conduct an ablation study to only leverage the content features for USE task. 
It means that the GRU only models the sequential information of the dialogue context without the direct effects of the DAR task. 
Table~\ref{ablation} shows that ``- DA Feat.'' performs poorly, indicating the importance of the signal provided by DA features in the USE task. 
Besides, we would also like to know about the effect of sequential modeling on USE task. We modify the USDA method by  removing the GRU layer and only employing a mean-pooling operation to aggregate the whole dialogue information, \textit{i.e.}, ``-GRU''. 
Results in Table~\ref{ablation} verify that either way hurts the performance of USE task, and thus the sequential modeling of the dialogue act and content is crucial to improve the performance of USE task.  
In addition, removing the CRF layer largely affects the DAR performance, which leads to a negative impact on USE task. 

\begin{table}
    \centering
    \caption{Ablation study (F1 scores on validation set).}
    \label{ablation}
    \begin{adjustbox}{max width=0.48\textwidth}
    \setlength{\tabcolsep}{1mm}{
    \begin{tabular}{llllll}
    \toprule
    \multirow{2}{*}{Model}&\multicolumn{2}{c}{USDA (MTL)}&\multicolumn{3}{c}{USDA (CLU)}\\
    \cmidrule(lr){2-3}\cmidrule(lr){4-6}
    &MWOZ&SGD&MWOZ&SGD&ReDial\\
    \midrule
    \textbf{USDA}&\textbf{49.7}&\textbf{60.2}&\textbf{48.4}&\textbf{58.3}&\textbf{57.1}\\
    \midrule
    - DA Feat.&47.9$_{-3.6\%}$&57.1$_{-5.1\%}$&47.0$_{-2.9\%}$&56.6$_{-2.9\%}$&55.3$_{-3.2\%}$\\
    - GRU&48.5$_{-2.4\%}$&58.7$_{-2.5\%}$&47.6$_{-1.7\%}$&57.6$_{-1.2\%}$&56.0$_{-1.9\%}$\\
    - CRF&48.4$_{-2.6\%}$&58.1$_{-3.5\%}$&-&-&-\\
    \midrule
    - DID&49.0$_{-1.4\%}$&58.8$_{-2.3\%}$&47.8$_{-1.2\%}$&58.1$_{-0.3\%}$&56.7$_{-0.7\%}$\\
    - SRS&48.8$_{-1.8\%}$&58.7$_{-2.5\%}$&47.0$_{-3.5\%}$&57.5$_{-1.4\%}$&56.6$_{-0.9\%}$\\
    - Pre-train&48.4$_{-2.6\%}$&57.3$_{-4.8\%}$&46.8$_{-3.3\%}$&57.8$_{-0.8\%}$&55.9$_{-2.1\%}$\\
    \bottomrule
    \end{tabular}}
    \end{adjustbox}
\end{table}

\subsubsection{Effect of Task-adaptive Pre-training}
The lower part in Table~\ref{ablation} reports the results of USDA with different task-adaptive pre-training strategies. 
Generally, both System Response Selection and Dialogue Incoherence Detection tasks contribute to the final performance more or less. 
In specific, pre-training with SRS task can bring more gains on the USE task, which verifies our motivation that the appropriateness of the system-provided response attaches great importance to the user satisfaction. 
Besides, the joint task-adaptive pre-training further enhances the performance by endowing the capability of both contextual and sequential dialogue modeling.

\subsection{Detailed Analyses \& Discussions}

\subsubsection{Quantitative Analysis of Feature Importance}
\begin{figure}
\centering
\includegraphics[width=0.48\textwidth]{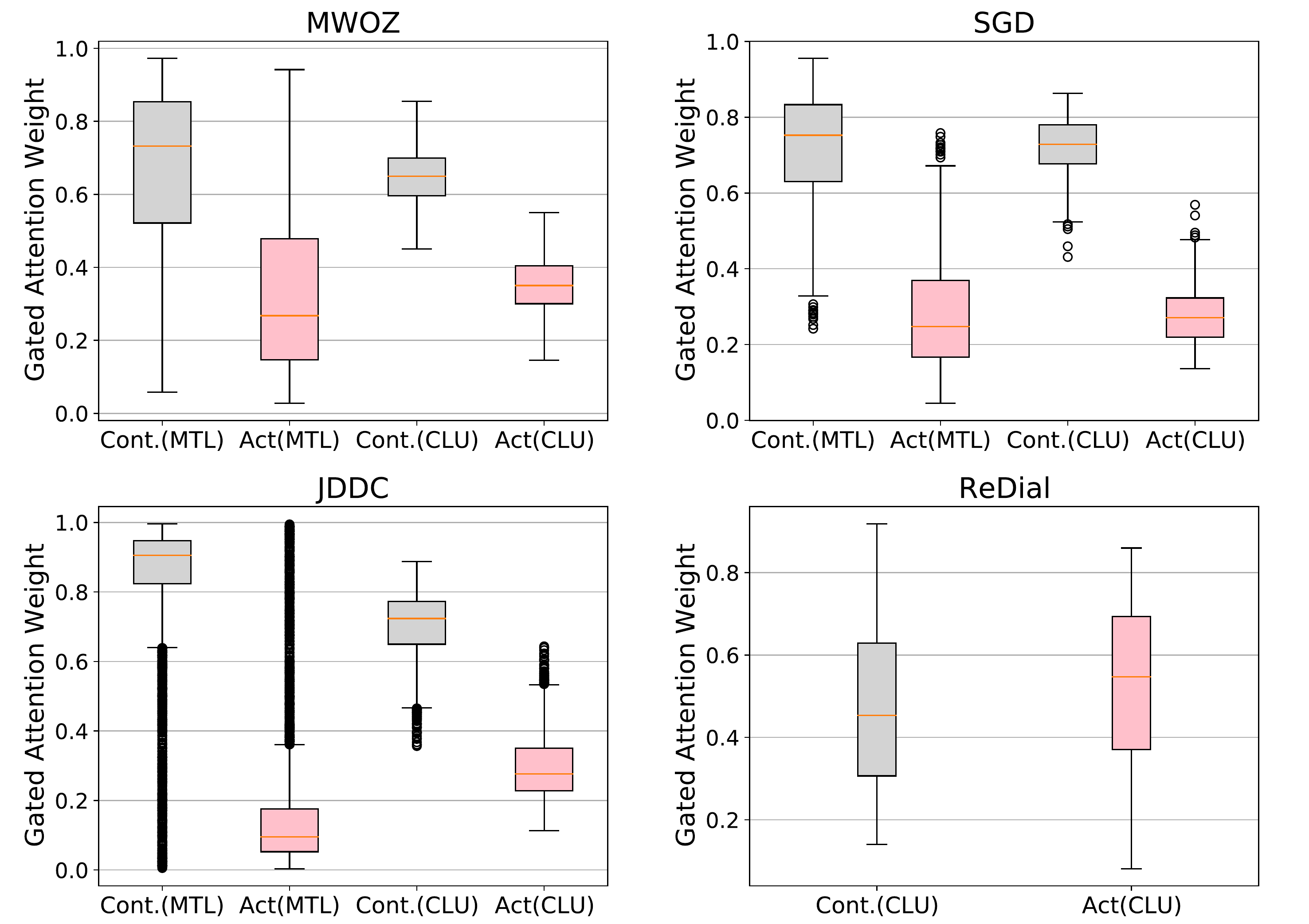}
\caption{Importance of Dialogue Content \& Act Features.}
\label{exp:weight}
\end{figure}

Figure~\ref{exp:weight} shows the distribution of the gated attention weight $g$ in Eq.~(\ref{eq:gate}), indicating the importance of content and DA features. 
Except for ReDial, the content features are generally assigned with higher weights than the DA features. The results show that the content features are still the dominant feature in most of the goal-oriented conversational systems for USE, while the DA features are more important in conversational recommendation, \textit{e.g.}, ReDial.  
Given the DA labels for supervised training (MTL), the gated attention weights of the DA features are more evenly distributed than those under unsupervised learning (CLU), indicating that the DA features are more discriminative with the prior knowledge from the predefined DA taxonomy. 
In JDDC dataset, the DA features learned by the supervised training make little contribution to USE task as the effect is restricted by the performance of DAR task, which also validates the overall performance of JDDC dataset in Section~\ref{sec:overall} (3). 

\subsubsection{Correlation between Predicted User Satisfaction and Dialogue Act Sequence}
\begin{figure}
\centering
\includegraphics[width=0.46\textwidth]{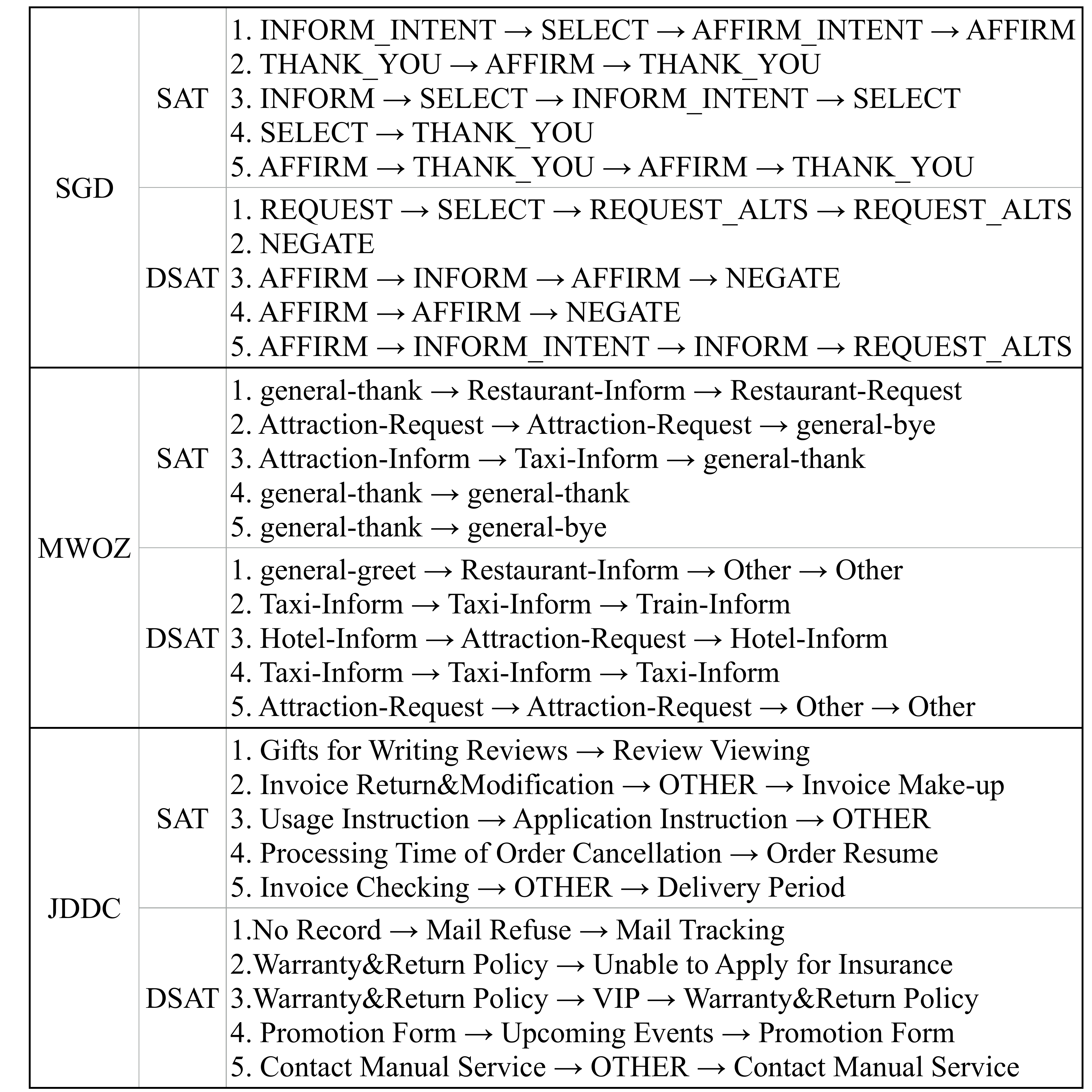}
\caption{Top Discriminative Dialogue Act Sub-sequences.}
\label{DAseq}
\end{figure}

To better understand the usefulness of specific DA sub-sequences in USE, we analyze the correlation between predicted DA sequences and predicted user satisfaction classes. 
Specifically, we define the following \textit{impact score} for a given DA sub-sequence $\bm{Q}$ towards the predicted satisfaction class $C$:   
\begin{equation}
    imp(\bm{Q},C) = \frac{1}{|\mathcal{S}_{\bm{Q}}|} \sum\nolimits_{\mathcal{D}\in\mathcal{S}_{\bm{Q}}^{C}} \left[(1-g)\frac{1}{|\bm{Q}|}\sum\nolimits_{y^d\in\bm{Q}} {\alpha_a}_{t:y^d}\right],
\end{equation}
where $\mathcal{S}_{\bm{Q}}$ and $\mathcal{S}_{\bm{Q}}^{C}$ denote the set of dialogues that contain the predicted DA sub-sequence $\bm{Q}$ and are predicted to be $C$. 
$g$ and $\alpha_a$ are the gated attention weight and the sequential DA weights obtained from Eq.~(\ref{eq:gate}) \& (\ref{eq:att}).  
The \textit{impact score} for a discriminative DA sub-sequence follows three intuitions: (i) The DA sub-sequence should appear more frequently in the concerned satisfaction class than others. (ii) The DA features should play a more important role in the cases that contain the DA sequence than content features. (iii) The attention weights of the DA sub-sequence in the whole dialogue should be higher. 
We present the top-5 discriminative predicted dialogue act sequences in Figure~\ref{DAseq}, which are the most influential in the predicted \textit{satisfied/dissatisfied} (SAT/DSAT) dialogues. 
This reveals the correlation between DA sequences and user satisfaction in different goal-oriented conversational systems. 
In addition, there is only one sub-sequence with a single DA, \textit{i.e.,} \textit{``NEGATE''}, among these most influential DA sub-sequences, implying the importance of considering sequential DA transitions in USE.

\section{Conclusions}
In this paper, we propose to leverage the sequential dynamics of dialogue acts to facilitate USE in goal-oriented conversational systems. A novel method, namely USDA, is proposed to jointly learn User Satisfaction Estimation and Dialogue Act Recognition tasks with a unified model, which supports both supervised and unsupervised DAR in terms of the availability of DA annotations. 
We further introduce two task-adaptive self-supervised pre-training strategies for enhancing the dialogue modeling capability of USDA. 
Extensive experiments show that USDA outperforms existing methods as well as validate the important role of dialogue act sequences in USE. 
USDA also provides a novel perspective to reveal the reason of user dis/satisfaction in goal-oriented conversational systems.

For future studies, it is worth exploring the causes of user dissatisfaction to better evaluate the goal-oriented conversational system. In real-world applications, the proposed simulation can raise interesting points to be further investigated with real users. 

\begin{acks}
This research/paper was supported by the Center for Perceptual and Interactive Intelligence (CPII) Ltd under the Innovation and Technology Commission's InnoHK scheme.
\end{acks}

%% file: appendix.tex
\appendix
\section{Appendix}
\subsection{Pretrain Data Preparation}\label{apx:pretrain}

As mentioned in Section~\ref{sec:pretrain}, the ``satisfactory'' samples in the original datasets are utilized as the positive samples for both System Response Selection (SRS) and Dialogue Incoherence Detection (DID) tasks. 

As for the negative samples for SRS task, we randomly pick $k$ system utterances in a positive sample, and each of them is replaced by a confounding utterance. 
The confounding utterance $u^*$ of a given utterance $u$ is retrieved from the entire corpus via BM25 under a threshold of 0.7. 
For example, the input of the positive sample is represented as $\{(u_{q_1},u_{p_1}),(u_{q_2},u_{p_2}),...,(u_{q_t},u_{p_t}),...\}$, and the SRS label will be $y^\text{SRS}=\{1,1,...,1,...\}^{T}$. 
If the system response at $t$-th exchange turn is replaced by $u^*$, the input of this negative sample will be $\{(u_{q_1},u_{p_1}),(u_{q_2},u_{p_2}),...,(u_{q_t},u^*),...\}$, and the SRS label will be $\{1,1,...,0,...\}^{T}$, which means that the system response at $t$-th exchange turn is inappropriate.
And $k$ is an integer randomly sampled from the range of [1,$T$/2]. 

As for the negative samples for DID task, we also randomly pick $k$ exchange turns in a positive sample, either remove them or shuffle them to construct the incoherent dialogue. 
For example, after we delete the second turn of exchange, the negative sample will be $\{(u_{q_1},u_{p_1}),(u_{q_3},u_{p_3}),...,(u_{q_t},u_{p_t}),...\}$.  
Or after we shuffle the first and the second turn of exchanges, the negative sample will be  $\{(u_{q_2},u_{p_2}),(u_{q_1},u_{p_1}),...,(u_{p_t},u_{q_t}),...\}$. 

As for the joint pre-training, the constructed negative samples in SRS task can also be regarded as incoherent dialogues for DID task, \textit{i.e.}, the DID labels for all the negative SRS samples are 0. 
As for the constructed DID samples, all the system responses are supposed to be appropriate, since each individual exchange turn is unchanged, \textit{i.e.}, the SRS labels for all DID samples are $\{1,1,...,1,...\}^{T}$.

\subsection{Implementation Details}\label{apx:impl}

For the Exchange-level BERT Encoder, we use BERT$_\text{base}$ pretrained weights\footnote{\url{https://github.com/huggingface/transformers}}, including BERT$_\text{base}$ Chinese for the JDDC dataset. 
For the Dialogue-level Transformer Encoder, we apply a 2-layer Transformer encoder with the input size and the hidden size as 768. The size of the Transformer FFN inner representation size is set to be 3072, and ReLU is used as the activation function. The learning rate and the dropout rate are set to be 2e-5 and 0.1, respectively. 
We train up to 20 epochs with mini-batch size 16, and select the best checkpoints based on the F1 score of USE task on the validation set.  
The dialogue act cluster number $K$ is set to be 20, which is tuned on the validation set. 
The hyper-parameters $\lambda_1, \lambda_2$, and $\lambda$ are empirically set to 1, 10, and 0.01 for balancing losses and the regularization term. 
The task-adaptive pretraining follows the same training configuration for 10 epochs. 
The F1 of the SRS and DID tasks achieves around 85\%-90\% and 60\%-70\% respectively.

\subsection{Further Analyses \& Discussions}\label{apx:disc}

\subsubsection{Per-class Performance}
Table~\ref{per-class} summarizes the per-class performance of different methods, where SAT/NEU/DSAT refer to ``satisfied/neutral/dissatisfied" respectively. 
For all the methods, we observe that user dissatisfaction in the dialogue is easier to be identified, as the performance on DSAT class is generally better than the other two classes. 
Previous studies~\cite{aaai21-handoff,sigdial21-dissat} find that the successful estimation of user dissatisfaction is more helpful to real-world applications. 
The performance improvement of the proposed methods comes from the DSAT class to a great extend in MWOZ and JDDC datasets.  

\begin{table}
\fontsize{8.5}{10}\selectfont
    \centering
    \caption{Performance w.r.t. Class (F1 score).}
    \label{per-class}
    \begin{adjustbox}{max width=0.48\textwidth}
    \setlength{\tabcolsep}{1mm}{
    \begin{tabular}{lccccccccc}
    \toprule
    \multirow{2}{*}{Model}&\multicolumn{3}{c}{MWOZ}&\multicolumn{3}{c}{SGD}&\multicolumn{3}{c}{JDDC}\\
    \cmidrule(lr){2-4}\cmidrule(lr){5-7}\cmidrule(lr){8-10}
    &SAT&NEU&DSAT&SAT&NEU&DSAT&SAT&NEU&DSAT\\
    \midrule
    BERT~\cite{bert}&40.2&47.2&50.3&53.4&47.6&60.1&53.4&59.0&66.1\\
    Co-GAT~\cite{aaai21-cogat}&38.3&48.1&52.5&56.4&46.3&63.8&54.1&60.4&68.5\\
    JointUSE~\cite{naacl16-hiattgru}&42.3&49.5&50.1&57.2&44.3&70.4&49.3&61.1&67.5\\
    \midrule
    USDA (CLU)&45.7&48.8&56.1&54.9&51.5&69.4&55.4&64.4&72.2\\
    USDA (MTL)&43.5&50.4&57.9&58.3&50.2&71.8&55.8&61.2&70.8\\
    \bottomrule
    \end{tabular}}
    \end{adjustbox}
\end{table}

\begin{figure}
\centering
\includegraphics[width=0.48\textwidth]{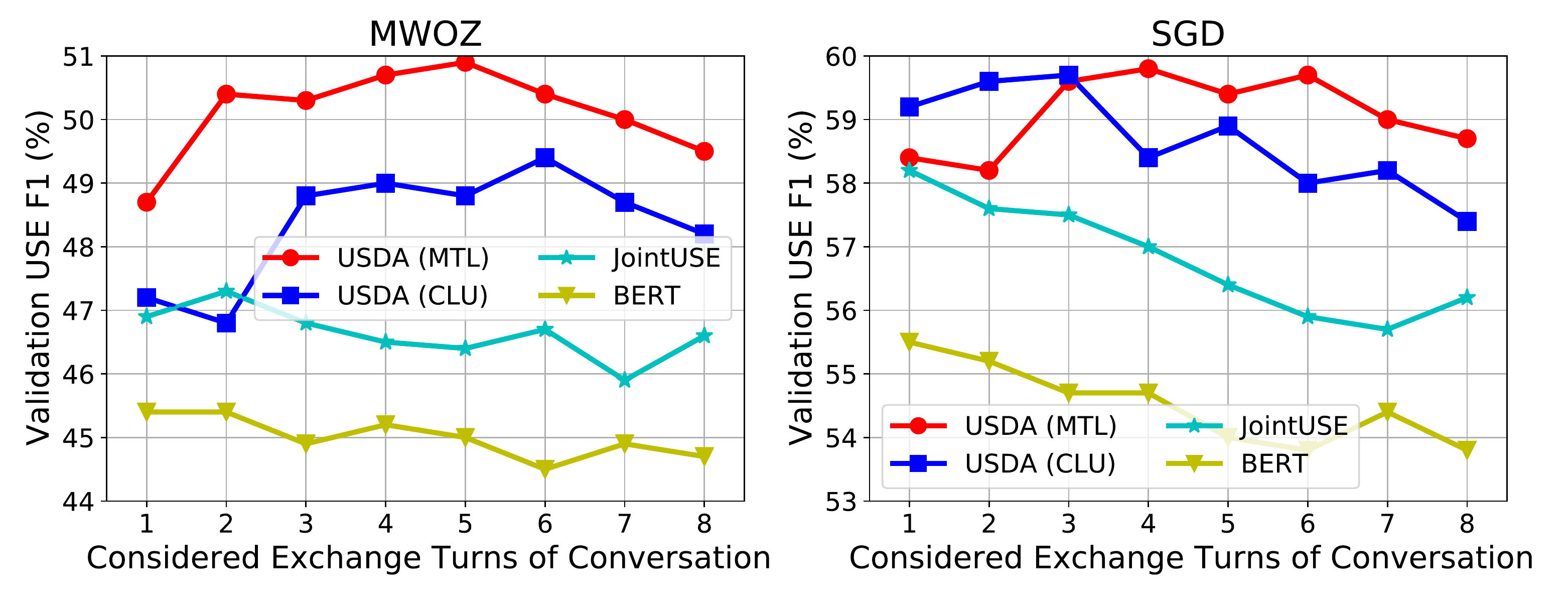}
\caption{Varying the Number of Dialogue Turns for USE.}
\label{exp:turn}
\end{figure}

\subsubsection{Varying the Number of Considered Dialogue Turns for USE}
Previous studies~\cite{umap20} find that the USE performance tends to decrease and fluctuate significantly when involving more conversation turns, since the investigated methods do not consider the sequential information of the dialogue.  
As shown in Figure~\ref{exp:turn}, we compare the proposed method with BERT~\cite{sigir21-uss} and JointUSE (+BERT)~\cite{emnlp20-jointuse}, by varying the number of considered dialogue turns. 
BERT preserves a relatively stable curve, since it takes the given conversation session as a whole to predict the user satisfaction and the input over 512 tokens will be truncated.  
The performance curve of JointUSE still drops with the increase of conversation turns, indicating JointUSE fails to handle the sequential information when encountering more turns of conversation. 
Conversely, USDA can reach a better performance when involving a certain number of conversation turns.